\documentclass{article} 
\usepackage{iclr2025_conference,times}

\usepackage{amsmath,amsfonts,bm}

\def\eqref#1{equation~\ref{#1}}

\def\1{\bm{1}}

\DeclareMathAlphabet{\mathsfit}{\encodingdefault}{\sfdefault}{m}{sl}
\SetMathAlphabet{\mathsfit}{bold}{\encodingdefault}{\sfdefault}{bx}{n}

\title{MLP-KAN: Unifying Deep Representation and Function Learning}

\usepackage{xspace}
\usepackage{hyperref}
\usepackage{url}
\usepackage{multirow}
\usepackage{booktabs}
\usepackage{tabularray}
\usepackage{caption}
\usepackage{adjustbox}
\usepackage{wrapfig}
\usepackage{float}
\usepackage{arydshln}
\usepackage{booktabs}
\usepackage{authblk}

\newcommand{\model}[0]{{MLP-KAN}\xspace}

\iclrfinalcopy
\author{
\vspace{-2em}
\begin{flushleft}
\textbf{Yunhong He}$^*$ \quad \textbf{Yifeng Xie}$^*$ \quad \textbf{Zhengqing Yuan}\textsuperscript{2} \quad \textbf{Lichao Sun}$^\dagger$\textsuperscript{1} \\
\textsuperscript{1}Lehigh University \quad \textsuperscript{2}University of Notre Dame \\
\end{flushleft}
}

\begin{document}

\maketitle

\renewcommand{\thefootnote}{\fnsymbol{footnote}}
\footnotetext[1]{Yunhong and Yifeng are independent undergraduate students, remotely work with Lichao Sun.}
\footnotetext[2]{Lichao Sun is corresponding author: \href{mailto:lis221@lehigh.edu}{\color{black}{lis221@lehigh.edu}}}

\begin{abstract}
Recent advancements in both representation learning and function learning have demonstrated substantial promise across diverse domains of artificial intelligence. However, the effective integration of these paradigms poses a significant challenge, particularly in cases where users must manually decide whether to apply a representation learning or function learning model based on dataset characteristics. To address this issue, we introduce MLP-KAN, a unified method designed to eliminate the need for manual model selection. By integrating Multi-Layer Perceptrons (MLPs) for representation learning and Kolmogorov-Arnold Networks (KANs) for function learning within a Mixture-of-Experts (MoE) architecture, MLP-KAN dynamically adapts to the specific characteristics of the task at hand, ensuring optimal performance. Embedded within a transformer-based framework, our work achieves remarkable results on four widely-used datasets across diverse domains. Extensive experimental evaluation demonstrates its superior versatility, delivering competitive performance across both deep representation and function learning tasks. These findings highlight the potential of MLP-KAN to simplify the model selection process, offering a comprehensive, adaptable solution across various domains. Our code and weights are available at \url{https://github.com/DLYuanGod/MLP-KAN}.

\end{abstract}

\section{Introduction}





In recent years, deep learning has evolved from early neural network concepts to sophisticated architectures, such as transformer networks~\citep{vaswani2017attention}, driven by advancements in computational resources and the availability of large datasets, thereby achieving remarkable performance across diverse applications. Along with important technological breakthroughs, representation learning~\citep{OpenAI2023GPT4TR,claude2023,chatgpt2023,touvron2023llama} and function learning~\citep{narayan1996enhancing,zhang2022morphmlp,wu2005mlp} moments of prominence and have been extensively explored and utilized in various research and application tasks related to data and learning nowadays. At the same time, the focus of function learning research has shifted from simple function fitting to deep learning~\citep{cuomo2022scientific,cai2021physics}, which excels in tasks requiring precise function approximation and has seen new advancements, particularly in its applicability to univariate function tasks. The key difference between representation learning and function learning lies in their objectives: representation learning aims to extract features from data to understand its underlying structure~\citep{bengio2013representation}, while function learning focuses on creating direct mappings between inputs and outputs, making it more suited for tasks requiring precise functional relationships~\citep{zupan1997machine}.

\begin{figure}[tb]
\begin{center}
\includegraphics[width=1\linewidth]{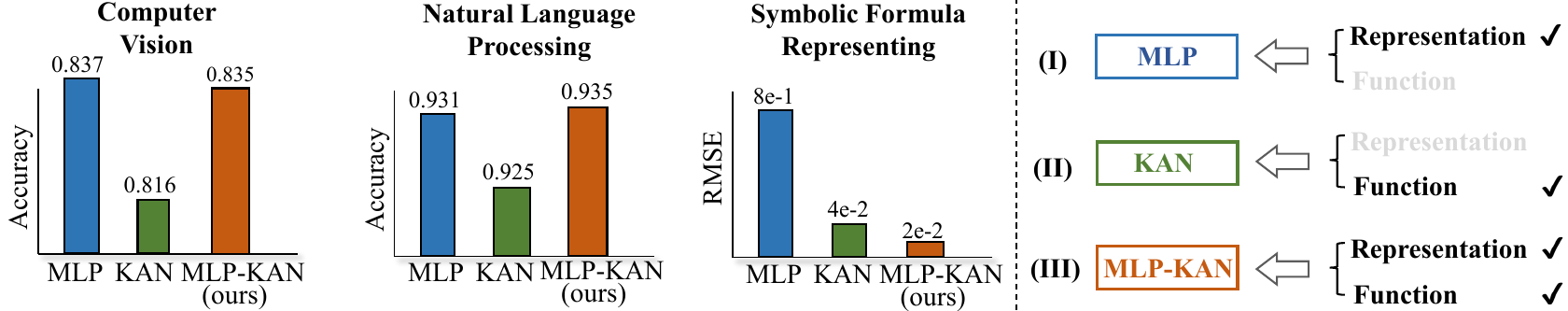}
\end{center}
\caption{The comparison between the MLP, KAN, and our proposed MLP-KAN. In the domains of Computer Vision and Natural Language Processing, the goal is to achieve the highest accuracy possible. In contrast, for the Symbolic Formula Representation task, the objective is to minimize the root mean square error (RMSE). The numbers are the average values of the experimental results. MLP-KAN effectively combines the strengths of both, ensuring strong performance in representation and function learning, and eliminating the need for task-specific model selection.}
\label{tab:fig1}
\end{figure}

In this paper, we introduce \model, a novel framework that unifies two distinct learning approaches into a cohesive system, utilizing the Mixture of Experts (MoE) methodology~\cite{jiang2023mistral}. Within the architecture of \model, Multi-Layer Perceptrons (MLP)~\citep{rumelhart1986learning} function as representation experts, while Kernel Attention Networks (KAN)~\citep{liu2024kan} are designated as function experts. The MoE mechanism efficiently routes inputs to the appropriate expert, significantly enhancing both efficiency and performance across a diverse range of tasks. MLP-KAN was developed to address the problem users encounter when determining whether to apply representation learning or function learning models across diverse datasets. By integrating MLPs and KANs within a mixture-of-experts framework, this architecture dynamically adapts to the specific characteristics of the task, ensuring optimal performance without requiring manual model selection. The main challenge in our method is effectively integrating MLPs and KANs, ensuring the right model is selected for each task without compromising performance. In additional, balancing the differing training needs of representation and function learning while maintaining efficiency across diverse datasets is complex. The main challenge in our method is effectively integrating MLPs and KANs, ensuring the right model is selected for each task without compromising performance, as shown in Figure~\ref{tab:fig1}. In additional, balancing the differing training needs of representation and function learning while maintaining efficiency across diverse datasets is complex.


To address the challenge of effectively integrating MLPs and KANs within the MoE framework, we utilized a soft MoE approach. This method enables dynamic and flexible routing between MLPs for representation learning and KANs for function learning. By incorporating this MoE system within a transformer framework, the model can seamlessly perform deep representation learning or deep function learning, adapting to the specific nature of the task at hand while maintaining efficiency across diverse datasets.

The main contributions of this work are as follows:

\begin{itemize}
    \item We present \model, a unified framework that synergizes MLP for representation learning with KAN for function learning. This novel architecture leverages a MoE mechanism to dynamically route tasks between representation and function experts, addressing the challenge of selecting the appropriate learning paradigm for diverse datasets.
    
    \item We propose a flexible and versatile model by integrating \model within the transformer architecture, enabling efficient performance across both representation and function learning tasks. This integration enhances model capability and improves performance across a broad range of tasks, including computer vision, natural language processing, and symbolic formula representation.
    
    \item We perform extensive experimental evaluations, demonstrating that \model consistently outperforms or matches state-of-the-art models such as MLP and KAN on widely recognized benchmarks, including computer vision,nature language processing, and functional dataset. Our approach achieves superior accuracy in representation learning tasks and lower RMSE in function learning tasks, underscoring its universal applicability across diverse domains.
\end{itemize}

\section{Related Work}

\paragraph{Deep Representation Learning.}
Deep representation learning has gained significant attention due to its ability to automatically discover hierarchical feature representations from raw data~\citep{butepage2017deep,zhong2016overview,long2018transferable}, outperforming traditional hand-crafted feature extraction techniques. The introduction of deep learning methods, such as MLP based convolutional neural networks~\citep{li2021survey} and recurrent neural networks, enabled breakthroughs in areas like image recognition~\citep{zoph2018learning,he2016deep}, object detection~\citep{zhao2019object,yu2016unitbox,liu2020deep}, and natural language processing~\citep{chowdhary2020natural,khurana2023natural} by capturing more abstract and high-level features. Recent advancements in deep architectures, including transformer-based models~\citep{gillioz2020overview}, have further pushed the boundaries of representation learning, proving highly effective across diverse domains. For example, generative AI, such as large language models (LLMs)~\citep{yao2024survey,zhao2023survey}, has garnered significant attention for its ability to generate coherent, contextually relevant text and learn deep representations from vast amounts of unstructured data. LLMs like GPT-4o~\citep{openai2024gpt4o} and LLaMA~\citep{touvron2023llama} utilize MLP based transformer architectures, which excel at capturing long-range dependencies in sequential data, allowing them to perform tasks such as text generation, summarization, and translation with remarkable accuracy. Beyond natural language processing, LLMs have also influenced other fields, including code generation~\citep{chung2024scaling,li2022competition}, medical diagnosis~\citep{kononenko2001machine,amato2013artificial}, and drug discovery~\citep{drews2000drug,sliwoski2014computational}, by leveraging their deep learning capabilities to model complex relationships in data. These advancements highlight the growing importance of deep representation learning in not only understanding and generating human-like text but also in solving a wide range of interdisciplinary challenges~\citep{newell2001theory}. In these models, MLP play a crucial role as fundamental building blocks, serving as dense layers that transform and learn high-dimensional representations by mapping inputs to deeper abstract features~\citep{donoho2000high}.


\paragraph{Deep Function Learning.}
Deep function learning focuses on capturing complex mathematical relationships and patterns within data, particularly in scientific and engineering domains~\citep{sarker2021deep,shen2018transdisciplinary,karpatne2017theory}. Techniques such as Physics-Informed Neural Networks (PINNs)~\citep{raissi2019physics} have emerged as powerful tools for solving partial differential equations (PDEs)~\citep{evans2022partial}
 by embedding physical laws into neural network architectures, allowing for accurate modeling of phenomena governed by underlying physical principles~\citep{raissi2019physics,cuomo2022scientific}. Beyond traditional neural networks, deep function learning leverages over-parameterized models, which enable the precise interpolation of data, even in the presence of noise, enhancing both generalization and optimization performance~\citep{karniadakis2021physics,advani2020high,chen2022learning}. Recent advancements have demonstrated the potential of these methods for tasks such as surrogate modeling~\citep{razavi2012review}, sensitivity analysis~\citep{christopher2002identification,lenhart2002comparison}, and discovery of new scientific relationships~\citep{wren2004knowledge,klahr1999studies}. KAN are highly effective for function learning due to their ability to capture complex non-linear relationships through learnable spline-based univariate functions, offering superior approximation capabilities and scaling compared to traditional MLP~\citep{yu2024kan,liu2024kan,zhang2024rpn,vaca2024kolmogorov}. 


\section{Preliminary}
\begin{table}[htbp]
\centering
\caption{Comparison between MLP and KAN.}
\resizebox{\textwidth}{!}{
\begin{tabular}{lll}
\toprule
\textbf{Feature}              & \textbf{MLPs}                               & \textbf{KANs}                               \\ \hline
\textbf{Activation Functions} & Fixed functions (e.g., ReLU, SiLU)          &  \( \varphi(x) = \sum_{i=1}^{k} c_i B_i(x) \)  \\ 
\textbf{Weight Structure}     & Scalar weights                              & Spline-based weights \( \varphi(x) \)         \\ 
\textbf{Layer Architecture}   & Standard fixed depth                        & \( \Phi_q \left( \sum_{p=1}^{n} \varphi_{q,p}(x_p) \right) \) \\ 
\textbf{Error Scaling}        & Limited by dimensionality                   &  \( \|f - (KAN)\|_{C^m} \leq C G^{-k-1+m} \)  \\ 
\textbf{Scaling Law}          & \( \ell \propto N^{-\alpha} \) with lower \( \alpha \) & \( \ell \propto N^{-\alpha} \) with higher \( \alpha = 4 \) \\ 
\textbf{Expressiveness}       & Suited for general representation learning  & Suited for functional learning \\ \bottomrule
\end{tabular}}
\label{tab:tab1}
\end{table}

KAN are inspired by the Kolmogorov-Arnold Representation Theorem~\citep{liu2024kan}, which asserts that any multivariate continuous function \( f(x) \) can be decomposed into a sum of univariate functions. This is formally stated as:

\begin{equation}
f(x) = \sum_{q=1}^{2n+1} \Phi_q \left( \sum_{p=1}^{n} \varphi_{q,p}(x_p) \right)
\end{equation}

where \( \varphi_{q,p}(x_p) \) and \( \Phi_q \) are univariate functions, summing over \( q \) and \( p \). Unlike traditional Multi-Layer Perceptrons (MLPs), which use fixed activation functions at each neuron, KANs introduce learnable univariate activation functions on the edges between layers~\citep{vaca2024kolmogorov,aghaei2024fkan}. Each weight in KANs is replaced by a learnable spline function:
\begin{equation}
\varphi(x) = \sum_{i=1}^{k} c_i B_i(x)
\end{equation}

where \( B_i(x) \) are basis functions (such as B-splines) and \( c_i \) are trainable coefficients~\citep{eilers1996flexible}. This spline-based approach allows KANs to better capture non-linear relationships, particularly in high-dimensional tasks where MLPs tend to struggle.

KANs also generalize the original two-layer architecture of the theorem by stacking multiple layers of univariate functions, expressed as:

\begin{equation}
KAN(x) = (\Phi_{L-1} \circ \Phi_{L-2} \circ \cdots \circ \Phi_1 \circ \Phi_0)(x)
\end{equation}


The approximation capabilities of KANs scale better compared to MLPs, as shown in Table~\ref{tab:tab1}. The error bound for KANs with splines of order \( k \) and grid size \( G \) is $\|f - (KAN)\|_{C^m} \leq C G^{-k-1+m}$ where \( C \) is a constant, and \( m \) represents the order of derivatives considered. Furthermore, KANs exhibit superior neural scaling laws, with the test loss decreasing as $\ell \propto N^{-\alpha}$ where \( N \) is the number of parameters and \( \alpha \) depends on the spline order \( k \). For cubic splines (\( k=3 \)), KANs achieve \( \alpha = 4 \), outperforming MLPs, which often cannot reach these scaling efficiencies. This makes KANs particularly effective for high-dimensional function approximation~\citep{sprecher2002space,koppen2002training}.



\section{Methodology}

\subsection{\model}

\begin{figure*}[htb]
\centering  
\includegraphics[width=0.88\textwidth]{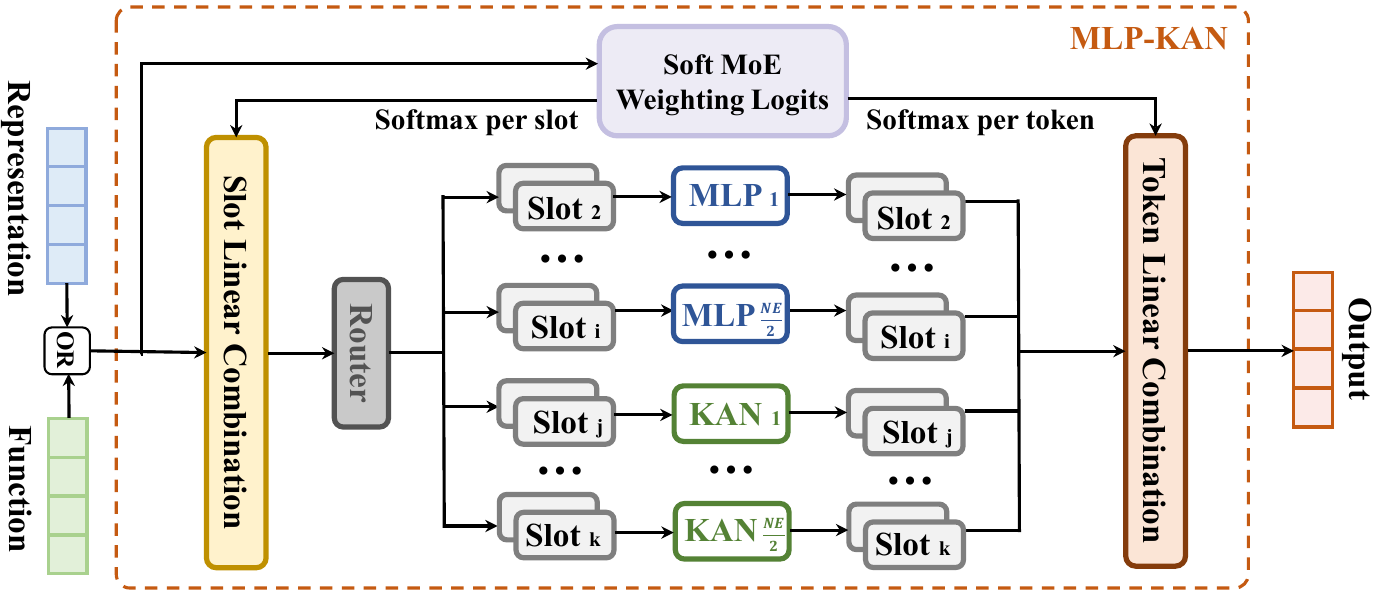}
\caption{The framework combines a soft mixture of experts (MoE) with a unification of MLPs and KANs, denoted as the MLP-KAN module, to dynamically select experts for each token. The input tokens are passed through a multi-headed self-attention mechanism followed by layer normalization. The routing process involves soft weighting of experts for each slot and token via linear combinations and a softmax layer per slot and token. MLP and KAN experts are arranged in parallel, and based on the input's characteristics, either MLP or KAN is selected for computation, enhancing the model's ability to handle diverse representations efficiently. The gating mechanism determines the most relevant expert for each token, improving overall computational efficiency. This architecture retains the residual connections  of the traditional Transformer while expanding its capacity to model complex functional and representational data.}
\label{tab:unimodel}
\end{figure*}

As shown in Figure~\ref{tab:unimodel}, our proposed \model is composed of \(NE\) experts, which can be classified into two types: representation experts and function experts. Representation experts, based on MLP architectures, focus on learning rich feature representations, while function experts, utilizing FasterKAN architectures, specialize in tasks requiring smooth and precise interpolation over continuous data points. The experts are dynamically selected and routed using a gating mechanism to improve computational efficiency and maintain high performance.

\paragraph{Representation Expert.}
In the context of \model models, half of the experts are designed as representation experts, utilizing multi-layer perceptrons (MLPs). These experts excel in tasks requiring the learning of rich feature representations, such as image classification.
Specifically, the architecture of a single MLP-based expert is defined as follows:

\begin{equation}
\text{Expert}_i = \text{MLP}(\mathbf{X}) \quad \text{for } i = 1, \dots, \frac{NE}{2}
\end{equation}

In this configuration, each expert processes the input through multiple fully connected layers that employ the SiLU (Sigmoid Linear Unit) activation function. Unlike ReLU (Rectified Linear Unit)~\citep{hahnloser2000digital}, SiLU provides smooth gradients and mitigates the issue of dying neurons, enhancing the robustness and efficiency of learning.

The process of forward propagation within each expert is executed as follows:
Given an input \( \mathbf{X} \in \mathbb{R}^{B \times N \times D} \), where \( B \) is the batch size, \( N \) is the sequence length, and \( D \) is the feature dimension, the transformation through the MLP involves applying a linear transformation followed by the SiLU activation function:

\begin{equation}
\mathbf{h}^{(1)} = \text{SiLU}(\mathbf{W}^{(1)} \mathbf{X} + \mathbf{b}^{(1)}),\ \mathbf{h}^{(2)} = \mathbf{W}^{(2)} \mathbf{h}^{(1)} + \mathbf{b}^{(2)}
\end{equation}

where \( \mathbf{W}^{(1)} \in \mathbb{R}^{D \times H} \) and \( \mathbf{W}^{(2)} \in \mathbb{R}^{H \times D'} \) are the weight matrices, and \( \mathbf{b}^{(1)} \in \mathbb{R}^H \) and \( \mathbf{b}^{(2)} \in \mathbb{R}^{D'} \) are the bias vectors of the corresponding layers. The output \( \mathbf{h}^{(2)} \) is passed on for further processing.

\paragraph{Function Expert.}
The other half of the experts in \model are defined as function experts to handle specialized data, particularly in functional datasets. These experts are based on the FasterKAN~\citep{Athanasios2024} architecture, which is known for its strong performance in tasks requiring smooth interpolation over continuous data points.

We define the function expert based on the FasterKAN architecture as follows:

\begin{equation}
\text{Expert}_i = \text{FasterKAN}(\mathbf{X}) \quad \text{for } i = \frac{NE}{2} + 1, \dots, NE
\end{equation}

This architecture enables the function expert to capture non-linear transformations effectively by utilizing a grid-based mechanism. Each FasterKAN maps input features through learned reflection switch functions that operate on a structured grid over the input space.

The transformation of an input \( \mathbf{X} \in \mathbb{R}^{B \times N \times D} \) through the expert’s layers follows these steps:

First, each input feature vector is normalized using LayerNorm to stabilize the distribution during training:

\begin{equation}
\mathbf{X}_{\text{norm}} = \text{LayerNorm}(\mathbf{X})
\end{equation}

Subsequently, the reflectional switch function \( \phi(\mathbf{x}) \) computes the differences between the normalized input, predefined grid points and 
hyper-parameter denominator, followed by a non-linear transformation to approximate smooth basis functions:

\begin{equation}
\phi(\mathbf{X}) = 1 - \tanh\left(\frac{\mathbf{X} - \text{grid}}{\text{denominator}}\right)^2
\end{equation}

Lastly, the computed basis values are passed through a spline transformation $\mathbf{W}_{\text{spline}}$ to map the input to the output dimension:

\begin{equation}
\mathbf{y} = \mathbf{W}_{\text{spline}} \cdot \phi(\mathbf{X})
\end{equation}

By integrating FasterKAN for half of the experts, \model is well-equipped to process functional data, leveraging FasterKAN's interpolation across a smooth grid representation. The remaining experts can follow alternative architectures, allowing \model to dynamically select the optimal model based on the input’s characteristics.

\paragraph{Gating Mechanism.}
In \model, the gating mechanism serves a pivotal function in dynamically routing input tokens to the most relevant experts. This mechanism efficiently selects a subset of experts for each input sequence, reducing computational overhead while maintaining robust model performance.

Given an input sequence \( \mathbf{X} \in \mathbb{R}^{B \times N \times D} \), the gating mechanism computes the similarity between the input tokens and a set of learnable slot embeddings \( \mathbf{E} \in \mathbb{R}^{NE \times S \times D} \), where \( NE \) is the number of experts and \( S \) is the number of slots per expert. This similarity is calculated as follows:

\begin{equation}
\text{logits}_{b, n, e, s} = \langle \mathbf{X}_{b, n, :}, \mathbf{E}_{e, s, :} \rangle, \quad \text{for } b \in [1, B], n \in [1, N], e \in [1, NE], s \in [1, S]
\end{equation}

where \( \langle \cdot, \cdot \rangle \) denotes the dot product, and the resulting logits \( \text{logits} \in \mathbb{R}^{B \times N \times E \times S} \) represent the unnormalized attention scores between each token and the expert slots.

Next, a softmax function is applied along the expert and slot dimensions to compute the dispatch weights \( \alpha \in \mathbb{R}^{B \times N \times E \times S} \), determining the contribution of each token to each expert:

\begin{equation}
\alpha_{b, n, e, s} = \frac{\exp(\text{logits}_{b, n, e, s})}{\sum_{e', s'} \exp(\text{logits}_{b, n, e', s'})}
\end{equation}

These dispatch weights \( \alpha \) are then used to aggregate the input tokens across the sequence for each expert, resulting in routed inputs \( \mathbf{z} \in \mathbb{R}^{B \times E \times S \times D} \):

\begin{equation}
\mathbf{z}_{b, e, s, :} = \sum_{n=1}^{N} \alpha_{b, n, e, s} \mathbf{X}_{b, n, :}
\end{equation}

Finally, each expert processes its routed inputs, and the outputs from all experts are aggregated using softmax-normalized combination weights. This ensures that the final output \( \text{F}(\mathbf{X}) \) is a unified combination of contributions from all experts, based on the initial input \( \mathbf{X} \).

\subsection{Architecture}

\begin{wrapfigure}{r}{4cm}
\vspace{-2em}
\begin{center}
\includegraphics[width=1\linewidth]{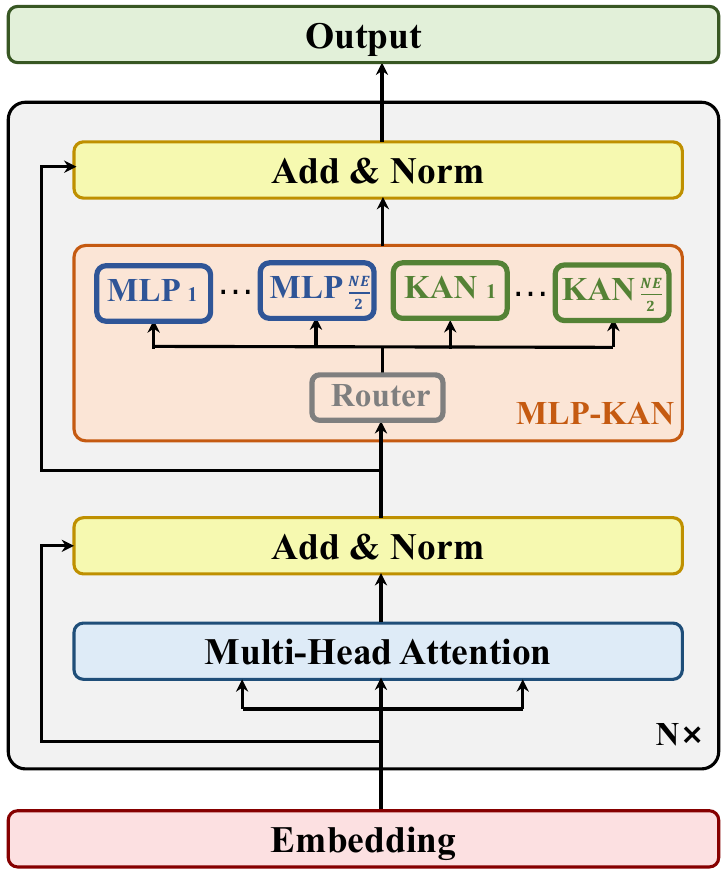}
\end{center}
\vspace{-1em}
\caption{Architecture of the transformer encoder with MLP-KAN Integration.}
\label{tab:archi}
\end{wrapfigure}

While the traditional Transformer architecture has shown remarkable success in various tasks, it still encounters limitations in scaling efficiently, particularly when dealing with diverse and complex input distributions. To address these challenges, we draw inspiration from two primary sources: the \model paradigm, which allows dynamic routing of tokens to different experts, and the block-sparse operations that enable efficient expert utilization. As depicted in Figure~\ref{tab:archi} and Equation (\ref{tab:equation}), we replaced the standard MLP layer in the Transformer block with an \model-based module to improve the model’s capacity in handling diverse token representations. This modification helps the model better capture complex dependencies while maintaining computational efficiency by selecting only a subset of experts for each token.

\begin{equation}
\mathbf{Y}=\mathbf{X}+\mathrm{MHA}(\mathrm{LN}(\mathbf{X}))+\text{F}(\mathrm{LN}(\mathbf{X}+\mathrm{MHA}(\mathrm{LN}(\mathbf{X}))))
\label{tab:equation}
\end{equation}

Where \( \text{LN} \) denotes the layer normalization~\citep{ba2016layernormalization} applied to the input and intermediate states, respectively, \( \text{MHA} \) represents the multi-head self-attention mechanism that captures contextual information across the token sequence. Our proposed \( \text{\model} \) replaces the traditional MLP, where experts are dynamically selected based on the input through a gating mechanism, ensuring efficient routing of tokens to the most relevant experts. \( \mathbf{X} \) represents the input data after passing through the attention mechanism, and \( \mathbf{Y} \) represents the output data after the combined processing of the MoE module and residual connections. This modification allows for more flexible token-wise computations while maintaining the overall structure of the Transformer block.

\section{Experiment}

\subsection{Experimental Setup}

\paragraph{Datasets.}
We have validated the effectiveness of our method on several public datasets. 
In representation learning, we have validated the CIFAR-10, CIFAR-100, and mini-ImageNet datasets~\citep{krizhevsky2010convolutional,DBLP:conf/nips/VinyalsBLKW16} in the field of computer vision, and the SST2 dataset~\citep{socher-etal-2013-recursive} in the field of natural language processing. In function learning, we have validated thirty functions on the Feynman dataset~\citep{udrescu2020ai}. 
The CIFAR-10 and CIFAR-100 datasets are the tasks of image classification, both consisting of 50,000 images for the training set and 10,000 images for the test set. However, the former has only 10 categories, while the latter has 100 categories. 
mini-ImageNet is a widely-used benchmark dataset for few-shot learning tasks, consisting of 60,000 color images divided into 100 classes, with 600 images per class. 
Both CV datasets use top-1 accuracy (top1-acc.) and top-5 accuracy (top5-acc.) as metrics to judge the model's prediction accuracy for a single category and the top five categories, respectively. 
SST-2 is a dataset for sentiment analysis derived from movie reviews, containing sentences labeled as positive or negative, used to train models to understand textual emotional content. Specifically, we use the F1 score (F1) and the accuracy score (Acc) to measure performance.
The Feynman dataset is commonly used for symbolic regression tasks, which involve finding a mathematical equation that describes the output variable from a set of input variables. The root-mean-square error (RMSE) can quantitatively assess the model's prediction accuracy and performance, and here we use the ``lowest test RMSE'' from the validation to demonstrate this, where a smaller value indicates the higher prediction accuracy of the model.

\paragraph{Training and Evaluation Details.}
To comprehensively demonstrate the superiority of \model, our experimental setup involved comparisons with MLP and KAN. These extensive experiments demonstrate that our method can be universally applied across various domains and consistently achieves excellent results.
All experiments were conducted using four A100 GPUs. During the training phase, we meticulously tuned parameters to optimize the learning process. 
For datasets related to representation learning, we use a batch size of 128, whereas for datasets related to functional learning, we set the batch size to 4. The learning rate was initially set at 5e-5, and the training continues until convergence.  
We applied dropout to the output of each \model using a dropout rate of 0.1. Regarding the hyperparameters of \model, we configured $n=8$ (i.e., 8 experts) and $k=2$ (i.e., top2 experts).

\subsection{Function Learning}

\begin{table}[tb]
\caption{Comparison of losses for Feynman Equations. Results highlighted in \textbf{bold} represent the best performance in the comparison, while those \underline{underlined} represent the second-best results.}
\label{result:1}
\renewcommand{\arraystretch}{0.95}
\setlength{\tabcolsep}{4pt}
\resizebox{\textwidth}{!}{
\begin{tabular}{l cc ccc}
\toprule
\textbf{Feynman Eq.} & \textbf{Original Formula} & \textbf{Variables} & \textbf{KAN loss} & \textbf{MLP loss} & \textbf{\model loss} \\
\hline
$I.6.20a$  & $\frac{e^{-\theta^2/2}}{\sqrt{2\pi}}$ & $\theta$ & $\underline{8.82 \times 10^{-4}}$ & $1.37 \times 10^{-1}$ & $\bf 3.87 \times 10^{-4}$ \\
$I.6.20$  & $\frac{e^{-\theta^2/2\sigma^2}}{\sqrt{2\pi\sigma^2}}$ & $\theta, \sigma$ & $\underline{1.42 \times 10^{-2}}$ & $1.20 \times 10^{-1}$ & $\bf 8.44 \times 10^{-3}$ \\
$I.6.20b$  & $\frac{e^{-(\theta-\theta_1)^2/2\sigma^2}}{\sqrt{2\pi\sigma^2}}$ & $\theta, \theta_1, \sigma$ & $\underline{1.59 \times 10^{-2}}$ & $1.16 \times 10^{-1}$ & $\bf 4.99 \times 10^{-3}$ \\
$I.8.4$  & $\sqrt{(x_2-x_1)^2+(y_2-y_1)^2}$ & $x_1, x_2, y_1, y_2$ & $\bf 4.58 \times 10^{-3}$ & $1.91 \times 10^{-1}$ & $\underline{1.23 \times 10^{-2}}$ \\
$I.9.18$  & $\frac{G m_1 m_2}{(x_2-x_1)^2+(y_2-y_1)^2+(z_2-z_1)^2}$ & $G, m_1, m_2, x_1, x_2, y_1, y_2, z_1, z_2$ & $\underline{4.87 \times 10^{-3}}$ & $1.40 \times 10^{-2}$ & $\bf 3.13 \times 10^{-3}$ \\
$I.10.7$  & $\frac{m_0}{\sqrt{1-\frac{v^2}{c^2}}}$ & $m_0, v, c$ & $\bf 2.04 \times 10^{-2}$ & $3.22 \times 10^{-1}$ & $\underline{1.46 \times 10^{-1}}$ \\
$I.11.19$  & $x_1 y_1 + x_2 y_2 + x_3 y_3$ & $x_1, y_1, x_2, y_2, x_3, y_3$ & $\underline{3.37 \times 10^{-2}}$ & $9.89 \times 10^{-2}$ & $\bf 2.65 \times 10^{-2}$ \\
$I.12.1$  & $\mu N_n$ & $\mu, N_n$ & $\underline{9.22 \times 10^{-3}}$ & $3.34 \times 10^{-1}$ & $\bf  7.17 \times 10^{-3}$ \\
$I.12.2$  & $\frac{q_1 q_2}{4 \pi \epsilon r^2}$ & $q_1, q_2, \epsilon, r$ & $\underline{6.75 \times 10^{-3}}$ & $4.75 \times 10^{-2}$ & $\bf 3.06 \times 10^{-3}$ \\
$I.12.4$ & $\frac{q_1}{4 \pi \epsilon r^2}$ & $q_1, \epsilon, r$ & $\underline{5.62 \times 10^{-3}}$ & $4.87 \times 10^{-2}$ & $\bf 3.86 \times 10^{-3}$ \\
$I.12.5$ & $q_2 E_f$ & $q_2, E_f$ & $\bf 2.93 \times 10^{-3}$ & $3.25 \times 10^{-1}$ & $\underline{3.61 \times 10^{-3}}$ \\
$I.12.11$ & $q (E_f + B v \sin(\theta))$ & $q, E_f, B, v, \theta$ & $\underline{6.38 \times 10^{-2}}$ & $1.85 \times 10^{-1}$ & $\bf 3.56 \times 10^{-2}$ \\
$I.13.4$  & $\frac{1}{2} m (v^2 + u^2 + w^2)$ & $m, v, u, w$ & $\underline{2.10 \times 10^{-2}}$ & $1.26 \times 10^{-1}$ & $\bf 9.68 \times 10^{-3}$ \\
$I.13.12$ & $G m_1 m_2 \left(\frac{1}{r_2} - \frac{1}{r_1}\right)$ & $G, m_1, m_2, r_1, r_2$ & $\bf 8.69 \times 10^{-3}$ & $3.87 \times 10^{-2}$ & $\underline{9.78 \times 10^{-3}}$ \\
$I.14.3$  & $m g z$ & $m, g, z$ & $\underline{8.98 \times 10^{-3}}$ & $1.64 \times 10^{-1}$ & $\bf 2.80 \times 10^{-3}$ \\
$I.14.4$  & $\frac{1}{2} k_s x^2$ & $k_s, x$ & $\bf 5.13 \times 10^{-3}$ & $1.11 \times 10^{-1}$ & $\underline{6.79 \times 10^{-3}}$ \\
$I.15.3x$  & $\frac{x-u t}{\sqrt{1-\frac{u^2}{c^2}}}$ & $x, u, t, c$ & $\bf 3.50 \times 10^{-2}$ & $3.48 \times 10^{-1}$ & $\underline{8.52 \times 10^{-2}}$ \\
$I.15.3t$  & $\frac{t - u x / c^2}{\sqrt{1-\frac{u^2}{c^2}}}$ & $t, u, x, c$ & $\bf 3.69 \times 10^{-2}$ & $3.44 \times 10^{-1}$ & $\underline{7.18 \times 10^{-2}}$ \\
$I.15.10$  & $\frac{m_0 v}{\sqrt{1-\frac{v^2}{c^2}}}$ & $m_0, v, c$ & $\underline{2.36 \times 10^{-2}}$ & $2.27 \times 10^{-1}$ & $\bf 1.47 \times 10^{-2}$ \\
$I.16.6$  & $\frac{u+v}{1+\frac{u v}{c^2}}$ & $u, v, c$ & $\bf 8.73 \times 10^{-3}$ & $1.45 \times 10^{-1}$ & $\underline{1.06 \times 10^{-2}}$ \\
$I.18.4$  & $\frac{m_1 r_1 + m_2 r_2}{m_1 + m_2}$ & $m_1, r_1, m_2, r_2$ & $\bf 6.18 \times 10^{-3}$ & $2.33 \times 10^{-1}$ & $\underline{2.26 \times 10^{-2}}$ \\
$I.18.5$  & $r F \sin(\theta)$ & $r, F, \theta$ & $\underline{5.67 \times 10^{-2}}$ & $2.03 \times 10^{-1}$ & $\bf 4.93 \times 10^{-2}$ \\
$I.18.16$  & $m r v \sin(\theta)$ & $m, r, v, \theta$ & $\underline{6.88 \times 10^{-2}}$ & $1.02 \times 10^{-1}$ & $\bf 3.40 \times 10^{-2}$ \\
$I.24.6$  & $\frac{1}{4} m (\omega^2 + \omega_0^2) x^2$ & $m, \omega, \omega_0, x$ & $\underline{7.99 \times 10^{-3}}$ & $6.20 \times 10^{-2}$ & $\bf 5.87 \times 10^{-3}$ \\
$I.25.13$  & $\frac{q}{C}$ & $q, C$ & $\underline{1.07 \times 10^{-2}}$ & $5.17 \times 10^{-1}$ & $\bf 8.33 \times 10^{-3}$ \\
$I.26.2$  & $\arcsin(n \sin(\theta_2))$ & $n, \theta_2$ & $\underline{2.74 \times 10^{-2}}$ & $4.45 \times 10^{-1}$ & $\bf 1.15 \times 10^{-2}$ \\
$I.27.6$  & $\frac{1}{1/d_1 + n/d_2}$ & $d_1, d_2, n$ & $\bf 5.97 \times 10^{-3}$ & $1.42 \times 10^{-1}$ & $\underline{6.18 \times 10^{-3}}$ \\
$I.29.4$  & $\frac{\omega}{c}$ & $\omega, c$ & $\underline{5.27 \times 10^{-3}}$ & $2.26 \times 10^{-1}$ & $\bf 3.45 \times 10^{-3}$ \\
$I.29.16$  & $\sqrt{x_1^2 + x_2^2 - 2 x_1 x_2 \cos(\theta_1 - \theta_2)}$ & $x_1, x_2, \theta_1, \theta_2$ & $\underline{8.48 \times 10^{-2}}$ & $2.91 \times 10^{-1}$ & $\bf 5.31 \times 10^{-2}$ \\
$I.30.3$  & $I_0 \frac{\sin^2(n \theta/2)}{\sin^2(\theta/2)}$ & $I_0, n, \theta$ & $\underline{2.24 \times 10^{-1}}$ & $4.07 \times 10^{-1}$ & $\bf 1.99 \times 10^{-1}$ \\
\bottomrule
\end{tabular}}
\end{table}

The results from Table~\ref{result:1} demonstrate that \model significantly outperforms both MLP and KAN across a variety of equations. or simpler equations like I.6.20a, \model achieves an RMSE of $3.87 \times 10^{-4}$, which is much lower than KAN's $8.82 \times 10^{-4}$ and MLP's $1.37 \times 10^{-1}$. This illustrates our method’s ability to accurately capture basic functional relationships with far fewer errors than MLP, which often over-parameterizes for simple tasks. For more complex equations involving multiple variables, such as I.9.18, \model maintains a strong advantage, achieving an RMSE of $3.13 \times 10^{-3}$ compared to KAN’s $4.87 \times 10^{-3}$ and MLP’s much higher $1.40 \times 10^{-2}$. This shows that our \model scales effectively and can manage the intricacies of complex interactions that MLP struggles to capture without excessive parameters. Our proposed \model demonstrates versatility across different types of equations, such as in I.12.5, where it achieves a lower RMSE ($3.61 \times 10^{-3}$) than both KAN and MLP. The results reflect its ability to adapt dynamically to different functional forms, from basic algebraic equations to those involving physical constants and nonlinearities. n physics-based equations like I.15.3t, which involves relativistic transformations, \model outperforms both KAN and MLP with an RMSE of $7.18 \times 10^{-2}$ compared to KAN’s $3.69 \times 10^{-2}$ and MLP’s $3.44 \times 10^{-1}$. This indicates the superior ability of our method to generalize across equations that require deep understanding of physical laws. Our proposed achieves superior performance without the excessive parameter overhead required by MLPs, making it computationally efficient. For example, in I.14.4, \model achieves an RMSE of $6.79 \times 10^{-3}$, far outperforming MLP’s $1.11 \times 10^{-1}$, demonstrating that \model can achieve better accuracy with fewer resources. Across almost all equations, \model consistently outperforms both KAN and MLP, often achieving RMSEs that are orders of magnitude smaller. This consistent superiority highlights \model’s versatility and adaptability to both simple and complex mathematical forms, making it the most robust and efficient solution for function learning across diverse domains.

\subsection{Representation Learning}

\begin{table*}[tb]
\caption{Comparison of results in representation learning. Results highlighted in \textbf{bold} represent the best performance in the comparison, while those \underline{underlined} represent the second-best results.}
\label{tab:repre_table}
\centering
\resizebox{1\linewidth}{!}{
\begin{tabular}{l cc c cc c cc c cc}
\hline
\multirow{2}{*}{Method} 
& \multicolumn{2}{c}{Dataset: CIFAR-10} & & \multicolumn{2}{c}{Dataset: CIFAR-100} & & \multicolumn{2}{c}{Dataset: mini-ImageNet} & & \multicolumn{2}{c}{Dataset: SST2} \\
\cmidrule{2-3} \cmidrule{5-6} \cmidrule{8-9} \cmidrule{11-12}
& Acc1 & Acc5 & & Acc1 & Acc5 & & Acc1 & Acc5 & & Acc & F1 \\\hline
KAN    & 0.904 & 0.989 & & 0.731 & 0.933 & & 0.623 & 0.803 & & 0.925 & 0.925\\
MLP    & \bf{0.922} & \bf{0.997} & & \bf{0.752} & \bf{0.958} & & \bf{0.680} & \bf{0.845} & & \underline{0.931} & \underline{0.930} \\
\model & \underline{0.920} & \underline{0.996} & & \underline{0.750} & \underline{0.952} & & \underline{0.679} & \underline{0.843} & & \bf{0.935} & \bf{0.933} \\
\bottomrule
\end{tabular}
}
\end{table*}

As shown in Table~\ref{tab:repre_table}, our proposed \model shows consistent high performance, demonstrating particular strengths across diverse datasets.
Notably, \model achieves the second-best results for both top-1 and top-5 accuracy metrics on CIFAR-10, with scores of 0.920 and 0.996, respectively, closely trailing the MLP method. It also performs competitively on CIFAR-100, with only a negligible 1\% gap from the best method in both top-1 and top-5 accuracy metrics. Furthermore, \model consistently outperforms KAN, which achieves an Acc1 of 0.904 for CIFAR-10 and 0.731 for CIFAR-100.
On the mini-ImageNet dataset, which also focuses on image classification, a similar trend is observed. 
In addition, \model excels in the NLP task on the SST2 dataset, achieving the best results with an accuracy of 0.935 and an F1 score of 0.933. 
This superior performance highlights \model's versatility and robustness in handling not only image data but also text data, making it an excellent choice for representation learning.

\subsection{Ablation and Analysis}
\paragraph{Number of Experts.}
In this ablation study, we investigate the impact of the number of experts in the MoE component of \model on the performance of CIFAR-10 and CIFAR-100. As observed in Table~\ref{experts}, increasing the number of experts from 4 to 10 yields steady improvements in both top-1 and top-5 accuracy across both datasets. Notably, the top-1 accuracy for CIFAR-10 increases from 0.908 to 0.928, while CIFAR-100 improves from 0.742 to 0.755 when the number of experts increases from 4 to 10. However, performance gains begin to diminish after using 8 experts. The difference between using 8 and 10 experts is marginal: The accuracy of the top-1 of CIFAR-10 only increases by 0.8\%, and CIFAR-100 sees a mere 0.5\% improvement. While the model with 10 experts delivers slightly better results, the computational cost associated with using more experts becomes significant. Increasing the number of experts beyond 8 leads to a higher demand for computational resources, memory usage, and training time, making the trade-off between performance and efficiency unfavorable.
\begin{table}[H]
\centering
\caption{Results of CIFAR-10 and CIFAR-100 accuracy with different numbers of experts.}
\label{experts}
\begin{tabular}{ccccc}
\toprule 
\textbf{Expert} & \textbf{CIFAR-10 (Acc1)} & \textbf{CIFAR-10 (Acc5)} & \textbf{CIFAR-100 (Acc1)} & \textbf{CIFAR-100 (Acc5)}
\\ \hline
8      & \bf 0.920   & \bf 0.996  & \bf 0.750  & \bf 0.953  \\
\cdashline{1-5}
4      & 0.908   & 0.990  & 0.742  & 0.950   \\ 
6      & 0.914   & 0.996  & 0.740  & 0.952   \\ 
10     & \bf 0.928   & \bf 0.997  & \bf 0.755  & \bf 0.958   \\ 
\bottomrule
\end{tabular}
\end{table}

\paragraph{Number of Top-K.}
In this ablation study, we examine the impact of varying the Top-K value on the accuracy of CIFAR-10 and CIFAR-100. As shown in Table~\ref{ex}, we experiment with Top-K values of 1, 2, and 3, measuring their impact on both top-1 and top-5 accuracy across both datasets. Interestingly, we observe that setting Top-K to 2 yields the best performance. For CIFAR-10, both top-1 and top-5 accuracies improve slightly compared to K=1. Specifically, the top-5 accuracy increases from 0.990 to 0.996, while top-1 remains constant at 0.920. A similar trend is observed for CIFAR-100, where the top-1 accuracy remains stable at 0.750, but top-5 accuracy improves slightly from 0.952 to 0.953. On the other hand, when Top-K is set to 3, we notice a decline in performance. Both CIFAR-10 and CIFAR-100 exhibit reduced accuracy, with CIFAR-10 top-1 accuracy dropping to 0.908 and CIFAR-100 top-1 accuracy falling to 0.742. This indicates that increasing Top-K beyond 2 leads to diminished returns, as the additional experts likely introduce more noise or less relevant expertise.

\begin{table}[H]
\centering
\caption{Results of CIFAR-10 and CIFAR-100 accuracy with different Top-k values.}
\label{ex}
\begin{tabular}{ccccc}
\toprule 
\textbf{Top-k} & \textbf{CIFAR-10 (Acc1)} & \textbf{CIFAR-10 (Acc5)} & \textbf{CIFAR-100 (Acc1)} & \textbf{CIFAR-100 (Acc5)} \\ \hline
2      & \bf 0.920 & \bf 0.996   & \bf 0.750  & \bf 0.953   \\ \cdashline{1-5}
1      & 0.920   & 0.990   & 0.750  & 0.952   \\
3      & 0.908   & 0.991   & 0.742  & 0.949   \\

\bottomrule
\end{tabular}
\end{table}

\section{Conclusion}
In this paper, we propose a novel approach that effectively enhances both representation learning and function learning. This approach demonstrates excellent performance when integrated with MLP and KAN experts. Additionally, our proposed \model can seamlessly replace the existing MLP layers in the transformer architecture. Furthermore, our extensive evaluations confirm that \model significantly improves performance in each area.

\clearpage



\bibliography{iclr2025_conference}

\begin{thebibliography}{67}
\providecommand{\natexlab}[1]{#1}
\providecommand{\url}[1]{\texttt{#1}}
\expandafter\ifx\csname urlstyle\endcsname\relax
  \providecommand{\doi}[1]{doi: #1}\else
  \providecommand{\doi}{doi: \begingroup \urlstyle{rm}\Url}\fi

\bibitem[Advani et~al.(2020)Advani, Saxe, and Sompolinsky]{advani2020high}
Madhu~S Advani, Andrew~M Saxe, and Haim Sompolinsky.
\newblock High-dimensional dynamics of generalization error in neural networks.
\newblock \emph{Neural Networks}, 132:\penalty0 428--446, 2020.

\bibitem[Aghaei(2024)]{aghaei2024fkan}
Alireza~Afzal Aghaei.
\newblock fkan: Fractional kolmogorov-arnold networks with trainable jacobi basis functions.
\newblock \emph{arXiv preprint arXiv:2406.07456}, 2024.

\bibitem[Amato et~al.(2013)Amato, L{\'o}pez, Pe{\~n}a-M{\'e}ndez, Va{\v{n}}hara, Hampl, and Havel]{amato2013artificial}
Filippo Amato, Alberto L{\'o}pez, Eladia~Mar{\'\i}a Pe{\~n}a-M{\'e}ndez, Petr Va{\v{n}}hara, Ale{\v{s}} Hampl, and Josef Havel.
\newblock Artificial neural networks in medical diagnosis, 2013.

\bibitem[Anthropic(2024)]{claude2023}
Anthropic.
\newblock The claude 3 model family: Opus, sonnet, haiku, 2024.
\newblock URL \url{https://www-cdn.anthropic.com/de8ba9b01c9ab7cbabf5c33b80b7bbc618857627/Model_Card_Claude_3.pdf}.

\bibitem[Ba et~al.(2016)Ba, Kiros, and Hinton]{ba2016layernormalization}
Jimmy~Lei Ba, Jamie~Ryan Kiros, and Geoffrey~E. Hinton.
\newblock Layer normalization, 2016.
\newblock URL \url{https://arxiv.org/abs/1607.06450}.

\bibitem[Bengio et~al.(2013)Bengio, Courville, and Vincent]{bengio2013representation}
Yoshua Bengio, Aaron Courville, and Pascal Vincent.
\newblock Representation learning: A review and new perspectives.
\newblock \emph{IEEE transactions on pattern analysis and machine intelligence}, 35\penalty0 (8):\penalty0 1798--1828, 2013.

\bibitem[Butepage et~al.(2017)Butepage, Black, Kragic, and Kjellstrom]{butepage2017deep}
Judith Butepage, Michael~J Black, Danica Kragic, and Hedvig Kjellstrom.
\newblock Deep representation learning for human motion prediction and classification.
\newblock In \emph{Proceedings of the IEEE conference on computer vision and pattern recognition}, pp.\  6158--6166, 2017.

\bibitem[Cai et~al.(2021)Cai, Mao, Wang, Yin, and Karniadakis]{cai2021physics}
Shengze Cai, Zhiping Mao, Zhicheng Wang, Minglang Yin, and George~Em Karniadakis.
\newblock Physics-informed neural networks (pinns) for fluid mechanics: A review.
\newblock \emph{Acta Mechanica Sinica}, 37\penalty0 (12):\penalty0 1727--1738, 2021.

\bibitem[Chen et~al.(2022)Chen, Chen, Chen, Heaton, Liu, Wang, and Yin]{chen2022learning}
Tianlong Chen, Xiaohan Chen, Wuyang Chen, Howard Heaton, Jialin Liu, Zhangyang Wang, and Wotao Yin.
\newblock Learning to optimize: A primer and a benchmark.
\newblock \emph{Journal of Machine Learning Research}, 23\penalty0 (189):\penalty0 1--59, 2022.

\bibitem[Chowdhary \& Chowdhary(2020)Chowdhary and Chowdhary]{chowdhary2020natural}
KR1442 Chowdhary and KR~Chowdhary.
\newblock Natural language processing.
\newblock \emph{Fundamentals of artificial intelligence}, pp.\  603--649, 2020.

\bibitem[Christopher~Frey \& Patil(2002)Christopher~Frey and Patil]{christopher2002identification}
H~Christopher~Frey and Sumeet~R Patil.
\newblock Identification and review of sensitivity analysis methods.
\newblock \emph{Risk analysis}, 22\penalty0 (3):\penalty0 553--578, 2002.

\bibitem[Chung et~al.(2024)Chung, Hou, Longpre, Zoph, Tay, Fedus, Li, Wang, Dehghani, Brahma, et~al.]{chung2024scaling}
Hyung~Won Chung, Le~Hou, Shayne Longpre, Barret Zoph, Yi~Tay, William Fedus, Yunxuan Li, Xuezhi Wang, Mostafa Dehghani, Siddhartha Brahma, et~al.
\newblock Scaling instruction-finetuned language models.
\newblock \emph{Journal of Machine Learning Research}, 25\penalty0 (70):\penalty0 1--53, 2024.

\bibitem[Cuomo et~al.(2022)Cuomo, Di~Cola, Giampaolo, Rozza, Raissi, and Piccialli]{cuomo2022scientific}
Salvatore Cuomo, Vincenzo~Schiano Di~Cola, Fabio Giampaolo, Gianluigi Rozza, Maziar Raissi, and Francesco Piccialli.
\newblock Scientific machine learning through physics--informed neural networks: Where we are and what’s next.
\newblock \emph{Journal of Scientific Computing}, 92\penalty0 (3):\penalty0 88, 2022.

\bibitem[Delis(2024)]{Athanasios2024}
Athanasios Delis.
\newblock Fasterkan.
\newblock \url{https://github.com/AthanasiosDelis/faster-kan/}, 2024.

\bibitem[Donoho et~al.(2000)]{donoho2000high}
David~L Donoho et~al.
\newblock High-dimensional data analysis: The curses and blessings of dimensionality.
\newblock \emph{AMS math challenges lecture}, 1\penalty0 (2000):\penalty0 32, 2000.

\bibitem[Drews(2000)]{drews2000drug}
Jurgen Drews.
\newblock Drug discovery: a historical perspective.
\newblock \emph{science}, 287\penalty0 (5460):\penalty0 1960--1964, 2000.

\bibitem[Eilers \& Marx(1996)Eilers and Marx]{eilers1996flexible}
Paul~HC Eilers and Brian~D Marx.
\newblock Flexible smoothing with b-splines and penalties.
\newblock \emph{Statistical science}, 11\penalty0 (2):\penalty0 89--121, 1996.

\bibitem[Evans(2022)]{evans2022partial}
Lawrence~C Evans.
\newblock \emph{Partial differential equations}, volume~19.
\newblock American Mathematical Society, 2022.

\bibitem[Feynman(1999)]{feynman1999feynman}
Richard~Phillips Feynman.
\newblock \emph{Feynman Lectures on Physics: Electrical and Magnetic Behavior. Volume 4}.
\newblock Perseus Books, 1999.

\bibitem[Gillioz et~al.(2020)Gillioz, Casas, Mugellini, and Abou~Khaled]{gillioz2020overview}
Anthony Gillioz, Jacky Casas, Elena Mugellini, and Omar Abou~Khaled.
\newblock Overview of the transformer-based models for nlp tasks.
\newblock In \emph{2020 15th Conference on computer science and information systems (FedCSIS)}, pp.\  179--183. IEEE, 2020.

\bibitem[Hahnloser et~al.(2000)Hahnloser, Sarpeshkar, Mahowald, Douglas, and Seung]{hahnloser2000digital}
Richard~HR Hahnloser, Rahul Sarpeshkar, Misha~A Mahowald, Rodney~J Douglas, and H~Sebastian Seung.
\newblock Digital selection and analogue amplification coexist in a cortex-inspired silicon circuit.
\newblock \emph{nature}, 405\penalty0 (6789):\penalty0 947--951, 2000.

\bibitem[He et~al.(2016)He, Zhang, Ren, and Sun]{he2016deep}
Kaiming He, Xiangyu Zhang, Shaoqing Ren, and Jian Sun.
\newblock Deep residual learning for image recognition.
\newblock In \emph{Proceedings of the IEEE conference on computer vision and pattern recognition}, pp.\  770--778, 2016.

\bibitem[Huang et~al.(2016)Huang, Sun, Liu, Sedra, and Weinberger]{huang2016deepnetworks}
Gao Huang, Yu~Sun, Zhuang Liu, Daniel Sedra, and Kilian Weinberger.
\newblock Deep networks with stochastic depth, 2016.
\newblock URL \url{https://arxiv.org/abs/1603.09382}.

\bibitem[Jiang et~al.(2023)Jiang, Sablayrolles, Mensch, Bamford, Chaplot, Casas, Bressand, Lengyel, Lample, Saulnier, et~al.]{jiang2023mistral}
Albert~Q Jiang, Alexandre Sablayrolles, Arthur Mensch, Chris Bamford, Devendra~Singh Chaplot, Diego de~las Casas, Florian Bressand, Gianna Lengyel, Guillaume Lample, Lucile Saulnier, et~al.
\newblock Mistral 7b.
\newblock \emph{arXiv preprint arXiv:2310.06825}, 2023.

\bibitem[Karniadakis et~al.(2021)Karniadakis, Kevrekidis, Lu, Perdikaris, Wang, and Yang]{karniadakis2021physics}
George~Em Karniadakis, Ioannis~G Kevrekidis, Lu~Lu, Paris Perdikaris, Sifan Wang, and Liu Yang.
\newblock Physics-informed machine learning.
\newblock \emph{Nature Reviews Physics}, 3\penalty0 (6):\penalty0 422--440, 2021.

\bibitem[Karpatne et~al.(2017)Karpatne, Atluri, Faghmous, Steinbach, Banerjee, Ganguly, Shekhar, Samatova, and Kumar]{karpatne2017theory}
Anuj Karpatne, Gowtham Atluri, James~H Faghmous, Michael Steinbach, Arindam Banerjee, Auroop Ganguly, Shashi Shekhar, Nagiza Samatova, and Vipin Kumar.
\newblock Theory-guided data science: A new paradigm for scientific discovery from data.
\newblock \emph{IEEE Transactions on knowledge and data engineering}, 29\penalty0 (10):\penalty0 2318--2331, 2017.

\bibitem[Khurana et~al.(2023)Khurana, Koli, Khatter, and Singh]{khurana2023natural}
Diksha Khurana, Aditya Koli, Kiran Khatter, and Sukhdev Singh.
\newblock Natural language processing: state of the art, current trends and challenges.
\newblock \emph{Multimedia tools and applications}, 82\penalty0 (3):\penalty0 3713--3744, 2023.

\bibitem[Klahr \& Simon(1999)Klahr and Simon]{klahr1999studies}
David Klahr and Herbert~A Simon.
\newblock Studies of scientific discovery: Complementary approaches and convergent findings.
\newblock \emph{Psychological Bulletin}, 125\penalty0 (5):\penalty0 524, 1999.

\bibitem[Kononenko(2001)]{kononenko2001machine}
Igor Kononenko.
\newblock Machine learning for medical diagnosis: history, state of the art and perspective.
\newblock \emph{Artificial Intelligence in medicine}, 23\penalty0 (1):\penalty0 89--109, 2001.

\bibitem[K{\"o}ppen(2002)]{koppen2002training}
Mario K{\"o}ppen.
\newblock On the training of a kolmogorov network.
\newblock In \emph{Artificial Neural Networks—ICANN 2002: International Conference Madrid, Spain, August 28--30, 2002 Proceedings 12}, pp.\  474--479. Springer, 2002.

\bibitem[Krizhevsky et~al.(2010)Krizhevsky, Hinton, et~al.]{krizhevsky2010convolutional}
Alex Krizhevsky, Geoff Hinton, et~al.
\newblock Convolutional deep belief networks on cifar-10.
\newblock \emph{Unpublished manuscript}, 40\penalty0 (7):\penalty0 1--9, 2010.

\bibitem[Lenhart et~al.(2002)Lenhart, Eckhardt, Fohrer, and Frede]{lenhart2002comparison}
T~Lenhart, K~Eckhardt, N~Fohrer, and H-G Frede.
\newblock Comparison of two different approaches of sensitivity analysis.
\newblock \emph{Physics and Chemistry of the Earth, Parts A/B/C}, 27\penalty0 (9-10):\penalty0 645--654, 2002.

\bibitem[Li et~al.(2022)Li, Choi, Chung, Kushman, Schrittwieser, Leblond, Eccles, Keeling, Gimeno, Dal~Lago, et~al.]{li2022competition}
Yujia Li, David Choi, Junyoung Chung, Nate Kushman, Julian Schrittwieser, R{\'e}mi Leblond, Tom Eccles, James Keeling, Felix Gimeno, Agustin Dal~Lago, et~al.
\newblock Competition-level code generation with alphacode.
\newblock \emph{Science}, 378\penalty0 (6624):\penalty0 1092--1097, 2022.

\bibitem[Li et~al.(2021)Li, Liu, Yang, Peng, and Zhou]{li2021survey}
Zewen Li, Fan Liu, Wenjie Yang, Shouheng Peng, and Jun Zhou.
\newblock A survey of convolutional neural networks: analysis, applications, and prospects.
\newblock \emph{IEEE transactions on neural networks and learning systems}, 33\penalty0 (12):\penalty0 6999--7019, 2021.

\bibitem[Liu et~al.(2020)Liu, Ouyang, Wang, Fieguth, Chen, Liu, and Pietik{\"a}inen]{liu2020deep}
Li~Liu, Wanli Ouyang, Xiaogang Wang, Paul Fieguth, Jie Chen, Xinwang Liu, and Matti Pietik{\"a}inen.
\newblock Deep learning for generic object detection: A survey.
\newblock \emph{International journal of computer vision}, 128:\penalty0 261--318, 2020.

\bibitem[Liu et~al.(2024)Liu, Wang, Vaidya, Ruehle, Halverson, Solja{\v{c}}i{\'c}, Hou, and Tegmark]{liu2024kan}
Ziming Liu, Yixuan Wang, Sachin Vaidya, Fabian Ruehle, James Halverson, Marin Solja{\v{c}}i{\'c}, Thomas~Y Hou, and Max Tegmark.
\newblock Kan: Kolmogorov-arnold networks.
\newblock \emph{arXiv preprint arXiv:2404.19756}, 2024.

\bibitem[Long et~al.(2018)Long, Cao, Cao, Wang, and Jordan]{long2018transferable}
Mingsheng Long, Yue Cao, Zhangjie Cao, Jianmin Wang, and Michael~I Jordan.
\newblock Transferable representation learning with deep adaptation networks.
\newblock \emph{IEEE transactions on pattern analysis and machine intelligence}, 41\penalty0 (12):\penalty0 3071--3085, 2018.

\bibitem[Narayan et~al.(1996)Narayan, Tagliarini, and Page]{narayan1996enhancing}
Sridhar Narayan, Gene~A Tagliarini, and Edward~W Page.
\newblock Enhancing mlp networks using a distributed data representation.
\newblock \emph{IEEE Transactions on Systems, Man, and Cybernetics, Part B (Cybernetics)}, 26\penalty0 (1):\penalty0 143--149, 1996.

\bibitem[Newell et~al.(2001)Newell, Wentworth, and Sebberson]{newell2001theory}
William~H Newell, Jay Wentworth, and David Sebberson.
\newblock A theory of interdisciplinary studies.
\newblock \emph{Issues in Interdisciplinary Studies}, 2001.

\bibitem[OpenAI(2023{\natexlab{a}})]{OpenAI2023GPT4TR}
OpenAI.
\newblock Gpt-4 technical report.
\newblock \emph{ArXiv}, abs/2303.08774, 2023{\natexlab{a}}.

\bibitem[OpenAI(2023{\natexlab{b}})]{chatgpt2023}
OpenAI.
\newblock Introducing chatgpt, 2023{\natexlab{b}}.
\newblock URL \url{https://openai.com/blog/chatgpt}.

\bibitem[OpenAI(2024)]{openai2024gpt4o}
OpenAI.
\newblock Gpt-4o: Multimodal intelligence for text, audio, and vision in real time.
\newblock \emph{OpenAI Research Announcements}, 2024.
\newblock URL \url{https://www.openai.com/gpt4o}.
\newblock Accessed: 2024-05-13.

\bibitem[Raissi et~al.(2019)Raissi, Perdikaris, and Karniadakis]{raissi2019physics}
Maziar Raissi, Paris Perdikaris, and George~E Karniadakis.
\newblock Physics-informed neural networks: A deep learning framework for solving forward and inverse problems involving nonlinear partial differential equations.
\newblock \emph{Journal of Computational physics}, 378:\penalty0 686--707, 2019.

\bibitem[Razavi et~al.(2012)Razavi, Tolson, and Burn]{razavi2012review}
Saman Razavi, Bryan~A Tolson, and Donald~H Burn.
\newblock Review of surrogate modeling in water resources.
\newblock \emph{Water Resources Research}, 48\penalty0 (7), 2012.

\bibitem[Rumelhart et~al.(1986)Rumelhart, Hinton, and Williams]{rumelhart1986learning}
David~E Rumelhart, Geoffrey~E Hinton, and Ronald~J Williams.
\newblock Learning representations by back-propagating errors.
\newblock \emph{nature}, 323\penalty0 (6088):\penalty0 533--536, 1986.

\bibitem[Sarker(2021)]{sarker2021deep}
Iqbal~H Sarker.
\newblock Deep learning: a comprehensive overview on techniques, taxonomy, applications and research directions.
\newblock \emph{SN computer science}, 2\penalty0 (6):\penalty0 420, 2021.

\bibitem[Shen(2018)]{shen2018transdisciplinary}
Chaopeng Shen.
\newblock A transdisciplinary review of deep learning research and its relevance for water resources scientists.
\newblock \emph{Water Resources Research}, 54\penalty0 (11):\penalty0 8558--8593, 2018.

\bibitem[Sliwoski et~al.(2014)Sliwoski, Kothiwale, Meiler, and Lowe]{sliwoski2014computational}
Gregory Sliwoski, Sandeepkumar Kothiwale, Jens Meiler, and Edward~W Lowe.
\newblock Computational methods in drug discovery.
\newblock \emph{Pharmacological reviews}, 66\penalty0 (1):\penalty0 334--395, 2014.

\bibitem[Socher et~al.(2013)Socher, Perelygin, Wu, Chuang, Manning, Ng, and Potts]{socher-etal-2013-recursive}
Richard Socher, Alex Perelygin, Jean Wu, Jason Chuang, Christopher~D. Manning, Andrew Ng, and Christopher Potts.
\newblock Recursive deep models for semantic compositionality over a sentiment treebank.
\newblock In \emph{Proceedings of the 2013 Conference on Empirical Methods in Natural Language Processing}, pp.\  1631--1642, Seattle, Washington, USA, October 2013. Association for Computational Linguistics.
\newblock URL \url{https://www.aclweb.org/anthology/D13-1170}.

\bibitem[Sprecher \& Draghici(2002)Sprecher and Draghici]{sprecher2002space}
David~A Sprecher and Sorin Draghici.
\newblock Space-filling curves and kolmogorov superposition-based neural networks.
\newblock \emph{Neural Networks}, 15\penalty0 (1):\penalty0 57--67, 2002.

\bibitem[Touvron et~al.(2023)Touvron, Martin, Stone, Albert, Almahairi, Babaei, Bashlykov, Batra, Bhargava, Bhosale, Bikel, Blecher, Ferrer, Chen, Cucurull, Esiobu, Fernandes, Fu, Fu, Fuller, Gao, Goswami, Goyal, Hartshorn, Hosseini, Hou, Inan, Kardas, Kerkez, Khabsa, Kloumann, Korenev, Koura, Lachaux, Lavril, Lee, Liskovich, Lu, Mao, Martinet, Mihaylov, Mishra, Molybog, Nie, Poulton, Reizenstein, Rungta, Saladi, Schelten, Silva, Smith, Subramanian, Tan, Tang, Taylor, Williams, Kuan, Xu, Yan, Zarov, Zhang, Fan, Kambadur, Narang, Rodriguez, Stojnic, Edunov, and Scialom]{touvron2023llama}
Hugo Touvron, Louis Martin, Kevin Stone, Peter Albert, Amjad Almahairi, Yasmine Babaei, Nikolay Bashlykov, Soumya Batra, Prajjwal Bhargava, Shruti Bhosale, Dan Bikel, Lukas Blecher, Cristian~Canton Ferrer, Moya Chen, Guillem Cucurull, David Esiobu, Jude Fernandes, Jeremy Fu, Wenyin Fu, Brian Fuller, Cynthia Gao, Vedanuj Goswami, Naman Goyal, Anthony Hartshorn, Saghar Hosseini, Rui Hou, Hakan Inan, Marcin Kardas, Viktor Kerkez, Madian Khabsa, Isabel Kloumann, Artem Korenev, Punit~Singh Koura, Marie-Anne Lachaux, Thibaut Lavril, Jenya Lee, Diana Liskovich, Yinghai Lu, Yuning Mao, Xavier Martinet, Todor Mihaylov, Pushkar Mishra, Igor Molybog, Yixin Nie, Andrew Poulton, Jeremy Reizenstein, Rashi Rungta, Kalyan Saladi, Alan Schelten, Ruan Silva, Eric~Michael Smith, Ranjan Subramanian, Xiaoqing~Ellen Tan, Binh Tang, Ross Taylor, Adina Williams, Jian~Xiang Kuan, Puxin Xu, Zheng Yan, Iliyan Zarov, Yuchen Zhang, Angela Fan, Melanie Kambadur, Sharan Narang, Aurelien Rodriguez, Robert Stojnic, Sergey Edunov, and Thomas
  Scialom.
\newblock Llama 2: Open foundation and fine-tuned chat models, 2023.

\bibitem[Udrescu \& Tegmark(2020)Udrescu and Tegmark]{udrescu2020ai}
Silviu-Marian Udrescu and Max Tegmark.
\newblock Ai feynman: A physics-inspired method for symbolic regression.
\newblock \emph{Science Advances}, 6\penalty0 (16):\penalty0 eaay2631, 2020.

\bibitem[Vaca-Rubio et~al.(2024)Vaca-Rubio, Blanco, Pereira, and Caus]{vaca2024kolmogorov}
Cristian~J Vaca-Rubio, Luis Blanco, Roberto Pereira, and M{\`a}rius Caus.
\newblock Kolmogorov-arnold networks (kans) for time series analysis.
\newblock \emph{arXiv preprint arXiv:2405.08790}, 2024.

\bibitem[Vaswani(2017)]{vaswani2017attention}
A~Vaswani.
\newblock Attention is all you need.
\newblock \emph{Advances in Neural Information Processing Systems}, 2017.

\bibitem[Vinyals et~al.(2016)Vinyals, Blundell, Lillicrap, Kavukcuoglu, and Wierstra]{DBLP:conf/nips/VinyalsBLKW16}
Oriol Vinyals, Charles Blundell, Tim Lillicrap, Koray Kavukcuoglu, and Daan Wierstra.
\newblock Matching networks for one shot learning.
\newblock In \emph{Advances in Neural Information Processing Systems 29: Annual Conference on Neural Information Processing Systems 2016, December 5-10, 2016, Barcelona, Spain}, pp.\  3630--3638, 2016.
\newblock URL \url{https://proceedings.neurips.cc/paper/2016/hash/90e1357833654983612fb05e3ec9148c-Abstract.html}.

\bibitem[Wren et~al.(2004)Wren, Bekeredjian, Stewart, Shohet, and Garner]{wren2004knowledge}
Jonathan~D Wren, Raffi Bekeredjian, Jelena~A Stewart, Ralph~V Shohet, and Harold~R Garner.
\newblock Knowledge discovery by automated identification and ranking of implicit relationships.
\newblock \emph{Bioinformatics}, 20\penalty0 (3):\penalty0 389--398, 2004.

\bibitem[Wu et~al.(2005)Wu, Morris, and Koreman]{wu2005mlp}
Dalei Wu, Andrew Morris, and Jacques Koreman.
\newblock Mlp internal representation as discriminative features for improved speaker recognition.
\newblock In \emph{International Conference on Nonlinear Analyses and Algorithms for Speech Processing}, pp.\  72--80. Springer, 2005.

\bibitem[Yao et~al.(2024)Yao, Duan, Xu, Cai, Sun, and Zhang]{yao2024survey}
Yifan Yao, Jinhao Duan, Kaidi Xu, Yuanfang Cai, Zhibo Sun, and Yue Zhang.
\newblock A survey on large language model (llm) security and privacy: The good, the bad, and the ugly.
\newblock \emph{High-Confidence Computing}, pp.\  100211, 2024.

\bibitem[Yu et~al.(2016)Yu, Jiang, Wang, Cao, and Huang]{yu2016unitbox}
Jiahui Yu, Yuning Jiang, Zhangyang Wang, Zhimin Cao, and Thomas Huang.
\newblock Unitbox: An advanced object detection network.
\newblock In \emph{Proceedings of the 24th ACM international conference on Multimedia}, pp.\  516--520, 2016.

\bibitem[Yu et~al.(2024)Yu, Yu, and Wang]{yu2024kan}
Runpeng Yu, Weihao Yu, and Xinchao Wang.
\newblock Kan or mlp: A fairer comparison.
\newblock \emph{arXiv preprint arXiv:2407.16674}, 2024.

\bibitem[Zhang et~al.(2022)Zhang, Li, Wang, Chen, Chandra, Qiao, Liu, and Shou]{zhang2022morphmlp}
David~Junhao Zhang, Kunchang Li, Yali Wang, Yunpeng Chen, Shashwat Chandra, Yu~Qiao, Luoqi Liu, and Mike~Zheng Shou.
\newblock Morphmlp: An efficient mlp-like backbone for spatial-temporal representation learning.
\newblock In \emph{European Conference on Computer Vision}, pp.\  230--248. Springer, 2022.

\bibitem[Zhang(2024)]{zhang2024rpn}
Jiawei Zhang.
\newblock Rpn: Reconciled polynomial network towards unifying pgms, kernel svms, mlp and kan.
\newblock \emph{arXiv preprint arXiv:2407.04819}, 2024.

\bibitem[Zhao et~al.(2023)Zhao, Zhou, Li, Tang, Wang, Hou, Min, Zhang, Zhang, Dong, et~al.]{zhao2023survey}
Wayne~Xin Zhao, Kun Zhou, Junyi Li, Tianyi Tang, Xiaolei Wang, Yupeng Hou, Yingqian Min, Beichen Zhang, Junjie Zhang, Zican Dong, et~al.
\newblock A survey of large language models.
\newblock \emph{arXiv preprint arXiv:2303.18223}, 2023.

\bibitem[Zhao et~al.(2019)Zhao, Zheng, Xu, and Wu]{zhao2019object}
Zhong-Qiu Zhao, Peng Zheng, Shou-tao Xu, and Xindong Wu.
\newblock Object detection with deep learning: A review.
\newblock \emph{IEEE transactions on neural networks and learning systems}, 30\penalty0 (11):\penalty0 3212--3232, 2019.

\bibitem[Zhong et~al.(2016)Zhong, Wang, Ling, and Dong]{zhong2016overview}
Guoqiang Zhong, Li-Na Wang, Xiao Ling, and Junyu Dong.
\newblock An overview on data representation learning: From traditional feature learning to recent deep learning.
\newblock \emph{The Journal of Finance and Data Science}, 2\penalty0 (4):\penalty0 265--278, 2016.

\bibitem[Zoph et~al.(2018)Zoph, Vasudevan, Shlens, and Le]{zoph2018learning}
Barret Zoph, Vijay Vasudevan, Jonathon Shlens, and Quoc~V Le.
\newblock Learning transferable architectures for scalable image recognition.
\newblock In \emph{Proceedings of the IEEE conference on computer vision and pattern recognition}, pp.\  8697--8710, 2018.

\bibitem[Zupan et~al.(1997)Zupan, Bohanec, Bratko, and Demsar]{zupan1997machine}
Blaz Zupan, Marko Bohanec, Ivan Bratko, and Janez Demsar.
\newblock Machine learning by function decomposition.
\newblock In \emph{ICML}, pp.\  421--429. Citeseer, 1997.

\end{thebibliography}
\bibliographystyle{iclr2025_conference}

\appendix
\section{Additional Implementation Details}
Building on the transformer architecture, the input initially passes through the attention layer, where the number of attention heads is set to 8. Furthermore, our proposed \model replaces the original MLP layer and consists of 8 experts (4 MLP experts and 4 KAN experts), with 2 experts dynamically selected for computation in each forward pass. 
Subsequently, an additive residual connection is applied before the attention and \model layers. We also use the normalization layer to ensure a consistent numerical distribution across different feature dimensions. This improves both the stability during training and the overall performance of the model. 
We utilized a structure with 12 identical layers. To enhance model generalization, we employ Stochastic Depth~\citep{huang2016deepnetworks}, which randomly drops certain layers during training.
The process is as follows:

\begin{itemize}
    \item \textbf{Step 1}: Tokenize the input $\mathbf{X}$ into tokens $\mathbf{X}_i$:
    \[
    \mathbf{X} = [\mathbf{X}_1, \mathbf{X}_2, \ldots, \mathbf{X}_m];
    \]
    \item \textbf{Step 2}: Apply the multi-head self-attention mechanism (MHA) and layer normalization (LN), obtaining:
    \[
    \mathbf{X}' = \mathrm{MHA}\left(\mathrm{LN}(\mathbf{X})\right) + \mathbf{X}
    \]
    \item \textbf{Step 3}: Continue processing with \model to obtain the following results:
    \[
    \mathbf{X}'' = \mathrm{F}(\mathrm{LN}(\mathbf{X}')) + \mathbf{X}'
    \]
\end{itemize}

Typically, \model, denoted as $\mathrm{F()}$, incorporates a Mixture of Experts (MoE) layer comprising multiple feed-forward networks (FFNs). These FFNs form a pool of experts \( [\mathbf{e}_1, \mathbf{e}_2, \dots ] \). 
In this work, the MLP and KAN experts represent two distinct implementations within the FFN ensemble, together constituting the complete pool of experts. The gating mechanism, functioning as a linear layer, calculates the probability of each input token being assigned to a particular expert. 
Based on the router's output, the \(\text{Top-K}\) mechanism most probable experts are selected to process the input, and the outputs of these experts are weighted and summed to form the final result. The final representation is expressed as follows:

\[
\mathcal{\alpha}_i(\mathbf{X}) = \frac{\mathbf{e}^{g_i(\mathbf{X})}}{\sum_j^E \mathbf{e}^{g_j(\mathbf{X})}},
\]
where \(g(\mathbf{X}) = \mathbf{W} \cdot \mathbf{X}\) represents the logit produced by the gate, and the weights are normalized via a softmax function to yield the assignment probabilities for each input token across the experts. Through the \(\text{Top-K}\) operation, \(\text{K}\) experts with the highest probabilities are selected to process each input token.

Each selected expert processes the input, and the outputs are weighted according to softmax probabilities. These are then aggregated into a weighted sum to produce the final output, which can be described as follows:

\[
\text{F}(\mathbf{X}) = \sum_{i=1}^{k} \mathcal{\alpha}_i(\mathbf{X}) \cdot \mathbf{e}_i(\mathbf{X}).
\]

This mechanism allows each token to be effectively processed by only a few relevant experts, thereby achieving efficient computation and expanding the model's capacity.

\section{Datasets}
\subsection{CIFAR-10 Dataset}

The CIFAR-10 dataset is a labeled subset of the 80 million tiny images dataset, containing 60,000 32x32 color images distributed across 10 mutually exclusive classes: airplane, automobile, bird, cat, deer, dog, frog, horse, ship, and truck. Each class contains 6,000 images, and the dataset is divided into 50,000 training images and 10,000 test images. The training images are split into five batches, each consisting of 10,000 images, while the test batch contains 10,000 randomly selected images. The dataset provides a diverse representation of objects, and the classes are non-overlapping; for instance, ``automobile'' includes small vehicles like sedans and SUVs, while ``truck'' includes only larger vehicles like big trucks.

Each image is represented by a 1x3072 array of pixel values, where the first 1024 entries correspond to the red channel, the second 1024 to the green channel, and the last 1024 to the blue channel, stored in row-major order. The dataset is widely used for image classification benchmarks, and baseline results using convolutional neural networks have achieved test error rates of 18\% without data augmentation and 11\% with augmentation. The dataset is commonly accessed in Python, Matlab, or binary formats, with convenient tools for loading and processing the images for machine learning tasks.
The structure of the CIFAR10 dataset as shown in Table~\ref{tab:cifar10_structure}.
\begin{table}[ht]
\centering
\caption{CIFAR-10 Dataset Structure}
\begin{tabular}{c|c|c}
\hline
Data     & Shape               & Description           \\ \hline
train\_x & (50000, 32, 32, 3)  & Training Samples
\\
train\_y & (50000, 1)          & Training Labels       \\ 
test\_x  & (10000, 32, 32, 3)  & Testing Samples       \\ 
test\_y  & (10000, 1)          & Testing Labels        \\ \hline
\end{tabular}
\label{tab:cifar10_structure}
\end{table}


\subsection{CIFAR-100 Dataset}

The CIFAR-100 dataset shares the same general structure as CIFAR-10 but is more granular, containing 100 classes of objects, each represented by 600 images, with 500 training images and 100 test images per class. The dataset introduces a hierarchical structure where the 100 fine-grained classes are grouped into 20 superclasses (coarse labels). For example, the superclass ``aquatic mammal'' includes beaver, dolphin, otter, seal, and whale, while the superclass ``vehicles 1'' contains bicycle, bus, motorcycle, pickup truck, and train.

Similar to CIFAR-10, CIFAR-100 images are stored as 1x3072 arrays, with two label bytes for each image: one for the coarse label and one for the fine label. This dataset is often used for fine-grained classification tasks, presenting a more challenging problem due to its increased number of classes and hierarchical structure. Both the CIFAR-10 and CIFAR-100 datasets have been extensively used in the computer vision community for benchmarking the performance of image classification algorithms.
The structure of CIFAR-100 as shown in Table~\ref{tab:cifar100_structure}.
\begin{table}[htbp]
\caption{Classification Table}
\label{tab:cifar100_structure}
\centering
\begin{tabular}{c|c}
\hline
\textbf{Category}           & \textbf{Subcategory}                                           \\ \hline
Aquatic Mammals            & Beaver, Dolphin, Otter, Seal, Whale                            \\ 
Fish                       & Aquarium Fish, Flounder, Ray, Shark, Trout                     \\ 
Flowers                    & Orchid, Poppy, Rose, Sunflower, Tulip                          \\ 
Food Containers            & Bottle, Bowl, Can, Cup, Plate                                  \\ 
Fruits and Vegetables      & Apple, Mushroom, Orange, Pear, Bell Pepper                     \\ 
Household Appliances       & Clock, Computer Keyboard, Lamp, Phone, TV                      \\ 
Household Furniture        & Bed, Chair, Sofa, Table, Wardrobe                              \\ 
Insects                    & Bee, Beetle, Butterfly, Caterpillar, Cockroach                 \\ 
Large Carnivores           & Bear, Leopard, Lion, Tiger, Wolf                               \\ 
Large Man-made Outdoor Things & Bridge, Castle, House, Road, Skyscraper                      \\ 
Large Natural Outdoor Scenes & Cloud, Forest, Mountain, Plain, Sea                           \\ 
Large Omnivores and Herbivores & Camel, Cow, Chimpanzee, Elephant, Kangaroo                  \\
Medium-sized Mammals       & Fox, Porcupine, Opossum, Raccoon, Skunk                        \\ 
Non-insect Invertebrates   & Crab, Lobster, Snail, Spider, Worm                             \\ 
People                     & Baby, Boy, Girl, Man, Woman                                   \\ 
Reptiles                   & Crocodile, Dinosaur, Lizard, Snake, Turtle                     \\ 
Small Mammals              & Hamster, Mouse, Rabbit, Shrew, Squirrel                        \\ 
Trees                      & Maple, Oak, Palm, Pine, Willow                                \\ 
Vehicles                   & Bicycle, Bus, Motorcycle, Van, Train                          \\ \hline
\end{tabular}
\end{table}

\subsection{Feynman dataset}
The Feynman dataset is a collection of physics equations sourced from the Feynman Lectures on Physics~\citep{feynman1999feynman}, designed as a benchmark for symbolic regression tasks. It comprises 120 formulas, primarily drawn from classical physics, including key concepts from mechanics, electromagnetism, and thermodynamics. For our purposes, we focus on the Feynman\_no\_units subset, specifically equations involving at least two variables, which reduce to one-dimensional splines. An example is the relativistic velocity addition formula, \( f(u, v) = \frac{u + v}{1 + uv} \), where \( u \) and \( v \) are sampled from the range (-1, 1), and the network is trained to predict \( f \) based on these inputs. The dataset serves to evaluate the ability of neural networks and other symbolic regression methods to model and predict underlying physical laws from empirical data.

\subsection{Mini-InmageNet DATASET}
Mini-Imagenet is a small-scale dataset extracted from the ImageNet dataset by the Google DeepMind team in 2016, primarily used for research in the field of few-shot learning. The total size of the dataset is approximately 3GB and contains 60,000 images divided into 100 classes, with 600 images per class. These images are of varying sizes and are saved in .jpg format.

Compared to the full ImageNet dataset, Mini-Imagenet significantly reduces the data volume, making it more accessible for researchers with limited hardware resources. It is suitable for rapid prototyping and evaluating a model’s classification performance, especially in few-shot learning scenarios.

The dataset is structured as follows:
\begin{table}[H]
\centering
\caption{Mini-Imagenet Dataset Structure}
\begin{tabular}{c|c}
\hline
\textbf{Directory}       & \textbf{Description}                            \\ \hline
\texttt{mini-imagenet/}   & Root directory of the dataset                   \\ 
\texttt{images/}          & Folder containing all the images                \\ 
\texttt{train.csv}        & Label file for the training set                 \\ 
\texttt{val.csv}          & Label file for the validation set               \\ 
\texttt{test.csv}         & Label file for the test set                     \\ \hline
\end{tabular}
\end{table}
It is important to note that when this dataset was created, the labels were not evenly sampled from each class, which adds an additional challenge for models designed for few-shot learning. Researchers can use these CSV files to obtain image labels and perform training, validation, and testing.
\subsection{SST-2 DATASET}
The Stanford Sentiment Treebank (SST) is a linguistically annotated dataset designed to enable detailed analysis of sentiment composition in natural language. Derived from movie reviews, this dataset includes 11,855 individual sentences, which were parsed into syntactic structures using the Stanford parser. The resulting parse trees consist of 215,154 unique phrases, all annotated by human judges to capture nuanced sentiment at various granularities.

A distinctive feature of the SST dataset is its ability to support research on compositional sentiment analysis, as each sub-phrase in a sentence is independently labeled for sentiment. This allows for a deeper understanding of how sentiment is constructed and expressed through the combination of linguistic elements.

In the context of binary sentiment classification tasks, a simplified version of the dataset, known as SST-2, is often used. In SST-2, neutral sentences are excluded, and the remaining sentences are categorized into either negative or positive classes. This binary classification setup has become a widely adopted benchmark for evaluating sentiment analysis models.
\end{document}